\documentclass[sigconf]{acmart}

\usepackage{booktabs} 

\usepackage{amsmath}
\usepackage{graphicx}
\usepackage{siunitx}
\usepackage[colorinlistoftodos]{todonotes}

\usepackage{graphicx}
\usepackage{subcaption}

\usepackage{fancyhdr}
\renewcommand\footnotetextcopyrightpermission[1]{} 

\begin{document}

\title{Speech Recognition: Key Word Spotting through Image Recognition}

\author{Sanjay Krishna Gouda}
\affiliation{%
}
\email{sgouda@ucsc.edu}

\author{Salil Kanetkar}
\affiliation{%
  }
\email{skanetka@ucsc.edu}

\author{Vrindavan Harrison}
\affiliation{%
}
\email{vharriso@ucsc.edu}

\author{Manfred K Warmuth}
\affiliation{}
\email{manfred@ucsc.edu}


\begin{abstract}
The problem of identifying voice commands has always been a challenge due to the presence of noise and variability in speed, pitch, etc. We will compare the efficacy of several neural network architectures for the speech recognition problem. In particular, we will build a model to determine whether a one second audio clip contains a particular word (out of a set of 10), an unknown word, or silence. The models to be implemented are a CNN recommended by the Tensorflow Speech Recognition tutorial, a low-latency CNN, and an adversarially trained CNN. 

The result is a demonstration of how to convert a problem in audio recognition to the better-studied domain of image classification, where the powerful techniques of convolutional neural networks are fully developed. 
Additionally, we demonstrate the applicability of the technique of Virtual Adversarial Training (VAT) to this problem domain, functioning as a powerful regularizer with promising potential future applications. 
\end{abstract}

\maketitle

\section{Introduction}
In this project, we will work on the task of voice command recognition, following the TensorFlow Speech Recognition Challenge\footnote{www.kaggle.com/c/tensorflow-speech-recognition-challenge} that is being carried out on Kaggle. We will be using the Speech Commands Dataset \cite{speechcommands} to train and evaluate our model. 

We implement a few different models that each address different aspects of our problem. One consideration in constructing a Deep Neural Network solution is the large demand for memory and computing capacity on the machine running the model. For these reasons we explore model implementations that operate in environments where memory and computation resources are limited, as well as unrestricted environments.
\begin{figure}[h!]
	\centering
    \includegraphics[width=0.55\linewidth]{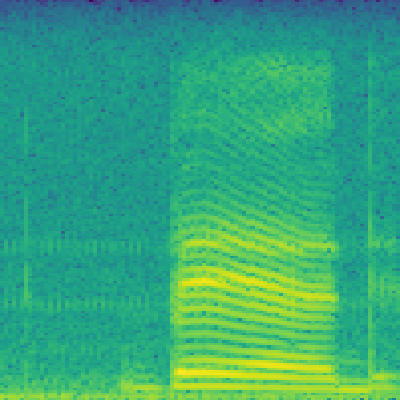}\\
    (a)\\
    \includegraphics[width=0.55\linewidth]{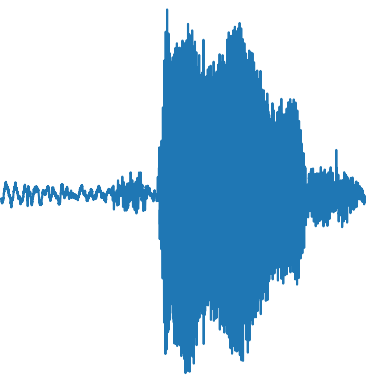}\\
    (b)\\
\caption{\label{fig:spectrogram}A comparison of the spectrogram (a) and the amplitude-vs.-time plot (b) for the same audio recording of a person saying the word ``bed''. }
\end{figure}
\section{Automatic Speech Recognition}
Speech Recognition is the sub-field of Natural Language Processing that focuses on understanding spoken natural language. This involves mapping auditory input to some word in a language vocabulary. The dataset we plan to work with has a relatively small vocabulary of 30 words. Our proposed model will learn to identify 10 of these 30 words, label any other words as unknown, and silence as silence. 

\subsection{Keyword Spotting}
Keyword spotting is a problem that was historically first defined in the context of speech processing. In speech processing, keyword spotting deals with the identification of keywords in utterances. This is especially useful in picking up on a finite list of command words, for example in voice command applications. 
Also, by picking out keywords from an utterance, a model can pick up on salient features and use them to get an understanding of the sentence as a whole. Humans perform a version of this task when interpreting hard-to-understand speech, such as an accent which is particularly fast or slurred, or a sentence in a language we do not know very well---we do not necessarily hear every single word that is said, but we pick up on salient key words and contextualize the rest to understand the sentence. 
Therefore key word spotting can be an important first step in picking up on the meaning of an utterance.
 
\subsection{Pre-processing}
One of the main aims of our project is to attempt to apply convolutional neural networks for speech recognition. 
Typical and well-studied CNN architectures are optimized for two-dimensional image processing tasks, an application where inputs have a very different shape than in audio processing: images are two-dimensional and not far from square (aspect ratios typically range from 1 to 2), so two-dimensional convolutions with square two-dimensional kernels are highly applicable. 
On the other hand, audio files are extremely long one-dimensional vectors. Our inputs, for example, were single-second audio files at a sample rate of \SI{16}{kHz}, so each training example was a vector in $\mathbb{R}^{16000}$. 
In image terms, this is equivalent to a one-pixel-high image with an aspect ratio of 16000, for which two-dimensional convolutions with square kernels are quite inapplicable.

While there is some precedent for using one-dimensional convolutional neural networks for audio processing (c.f. \cite{manganaro1999one}, whose abstract even begins by noting that CNNs are usually two-dimensional), we opted instead to modify the input data in order to make well-studied image processing techniques more applicable, by translating the input audio data into spectrograms.
A spectrogram is a method of using discrete Fourier transforms to translate a one-dimensional time sequence, most frequently but not necessarily a segment of audio, into a two-dimensional image containing the same information, but organized in a way that makes locality meaningful in two dimensions rather than just one.
Each column of the image represents the magnitude of the Fourier transform of the time sequence taken in a different contiguous window of time; the image as a whole represents a two-dimensional function from time and frequency to intensity rather than a one-dimensional function from time to absolute amplitude. 
Thus, in the resulting image, it is possible to directly see which frequency components change and how the frequency spectrum of the audio changes with time.

One would expect intuitively that this input format would be more conducive to analysis because the relevant information is directly visible; a comparison of the two types of sequence can be seen in \autoref{fig:spectrogram}.
While both figures show that there is a single loud section towards the end, only the spectrogram also shows the pitch contour of the word: each of the horizontal lines represents a single pitch component, and they all point somewhat downward. 
Additionally, there is biological precedent for the use of this input format: the human cochlea has different regions which are responsive to different frequencies. The result is that the input to the brain from the auditory sensory organ at any given time is comparable to a single column of the spectrogram: there are hundreds of different input neurons, each of whose firing rate is proportional to the strength of a particular frequency component. 

\section{Low Latency Convolutional Model }
Over the last decade the popularity of mobile devices has been increasing. Many applications that run on these devices use machine learning models. For example, it is now standard for new mobile devices to provide a speech-to-text service. Unfortunately, many machine learning models---especially deep neural networks---require heavy use of computation and memory resources that are lacking on mobile devices. In this section we describe a Low Latency Convolutional Neural Network that is designed to reduce its memory footprint by limiting the number of model parameters.  

The model described in this section is similar to the model called ``\texttt{cnn-one-fstride4}'' in \cite{Sainath}. 
(We based our implementation of this model on a script distributed by the TensorFlow-affiliated authors of the Kaggle challenge.) 
However, there are some small differences between our model and cnn-one-fstride4 that arise from how we preprocessed the audio inputs. The differences are shown in Table \ref{table:lowconv-conv-design}.

\begin{table}[h!]
\centering
\begin{tabular}{|l|c|c|c|c|c|}
	\hline
Model & Filter size & Stride & Channels & Dense & Params\\ 
\hline
(a)		& 8 &   4	& 1	 & 128 & 47.6k\\
(b) 	& 7 &   3	& 3  & 100 & 63.8k\\

\hline
\end{tabular}
\caption{Design of each model. (a) cnn-one-fstride4 described in \cite{Sainath}. (b) our model. }
\label{table:lowconv-conv-design}
\end{table}

\subsection{Model Design}
\begin{figure}[h!]
	\centering
    \frame{\includegraphics[height=100mm]{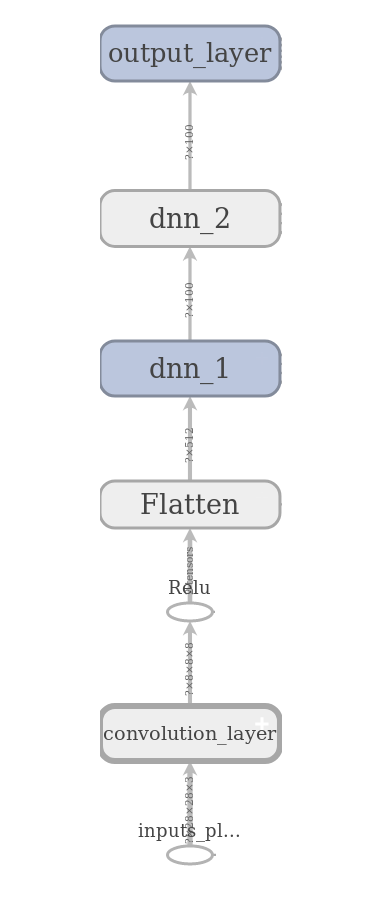}}
    \caption{A view of the computation graph for the low computation convolutional model.}
    \label{fig:lowconv_model_graph}
\end{figure}

The low latency model consists of a single convolutional layer that feeds into a Deep Neural Network. 
A block diagram of this model is shown in Figure \ref{fig:lowconv_model_graph}. A high level description of our network structure is as follows:
\begin{enumerate} 
	\setlength{\itemsep}{0pt}
    \setlength{\parskip}{0pt}
    \setlength{\parsep}{0pt} 
	\item convolution layer with nonlinearity
    \item fully connected feed forward layer without nonlinearity
    \item fully connected feed forward layer with nonlinearity 
    \item softmax output layer
\end{enumerate}
\noindent
The rest of this section will discuss the details of each layer.

\subsection{Variable Initialization}
Since training the CNN and DNN using Stochastic Gradient Descent is a non-convex optimization problem, parameter initialization is important.

We use Xavier Initialization to initialize our model weights. Xavier Initialization is described as follows: For an $m \times n$ dimensional matrix $\boldsymbol M$, $M_{ij}$ is assigned values selected uniformly from the distribution [$-\epsilon$,$\epsilon$], where
\begin{equation*}
\epsilon = \frac{\sqrt{6}}{\sqrt{m + n}}
\end{equation*}

We also initialize variables using a truncated normal distribution. In this initialization method, values are first selected from a normal distribution. Then any values that are over two standard deviations from the mean are rejected and re-drawn from the distribution. We sample values from a normal distribution that has a standard deviation of 0.01. 

\section{MNIST Tensorflow CNN }
We use the MNIST sample CNN architecture \textbf{[citation needed]} for this process.It can be found at github repository of tensorflow.   We will have to do some minor tweaks in the CNN's first layer so as not to get any dimension errors. 

\subsection{Model Design}
Its a deep NN with two convolution layers and two maxpool layers. The structure of the CNN is as follows:

\begin{enumerate} 
	\setlength{\itemsep}{0pt}
    \setlength{\parskip}{0pt}
    \setlength{\parsep}{0pt} 
	\item First Convolutional Layer
    \item Second Convolutional Layer
    \item Densely Connected Layer
    \item Softmax output layer
\end{enumerate}

\begin{figure}[h!]
	\centering
    \frame{\includegraphics[width=0.7\linewidth]{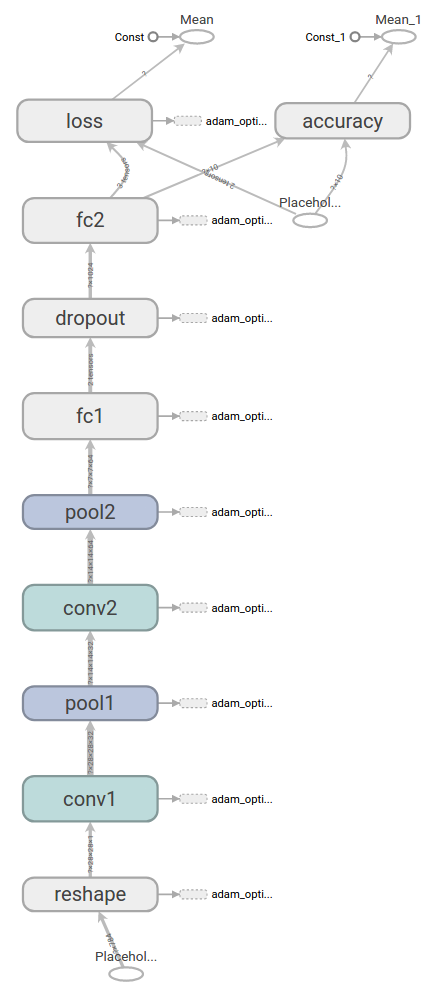}}
    \caption{A view of the computation graph for the MNIST model.}
    \label{fig:mnist_model_graph}
\end{figure}

\noindent
A diagram of this model's architecture is shown in Figure \ref{fig:mnist_model_graph}. The rest of this section will discuss the details of each layer.
\subsubsection{First Convolutional Layer}
The first layer consists of convolution, followed by max pooling. The convolution will compute 32 features for each 5x5 patch. Its weight tensor will have a shape of [5, 5, 1, 32]. The first two dimensions are the patch size, the next is the number of input channels, and the last is the number of output channels. We will also have a bias vector with a component for each output channel.

\subsubsection{Second Convolutional Layer}
In order to build a deep network, we stack several layers of this type. The second layer will have 64 features for each 5x5 patch.

\subsubsection{Densely Connected Layer}
Now that the image size has been reduced to 7x7, we add a fully-connected layer with 1024 neurons to allow processing on the entire image. We reshape the tensor from the pooling layer into a batch of vectors, multiply by a weight matrix, add a bias, and apply a ReLU. To avoid overfitting we also apply dropout.

\subsection{Base Model CNN}
A baseline architecture is described in \cite{Sainath}.

An additional variable to consider is the difference between time- and frequency-domain inputs. While one approach to the problem of improving non-locality in the network is to either add an attentional mechanism to an RNN or use dilated convolutions in a CNN, it is also possible to pre-process the data in order to perform learning in the frequency domain.
This has a precedent in the best known speech recognition system, the human ear, which in effect performs frequency analysis as a preprocessing step.
\begin{figure}[h!]
\frame{\includegraphics[width = 0.9\linewidth]{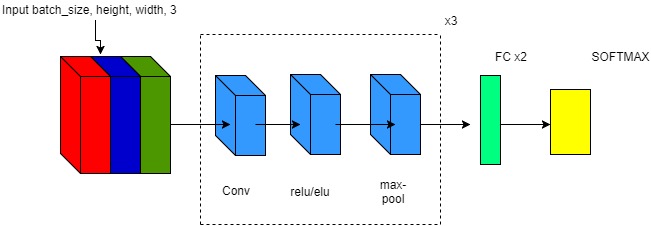}}
\caption{Shallow CNN Architecture}
\label{fig: shallow CNN}
\end{figure}
\section{CNNs and Virtual Adversarial Training}
\subsection{Architecture}
We have also tried a different architecture in CNNs. We followed MCDNN \cite{ciregan2012multi} and AlexNet \cite{krizhevsky2012imagenet} closely in having a sequence of convolutional layers followed by non linear activations and max-pooling and eventually flattening the convolved outputs into two fully connected layers.

\begin{figure}[h!]
\frame{\includegraphics[width = 0.9\linewidth]{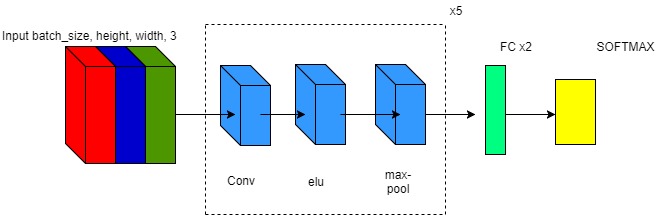}}
\caption{Deep CNN Architecture}
\label{fig:Wider CNN}
\end{figure}
Our first CNN model has a 5 layers where the first 3 layers are the sequence of conv-relu-maxpool (C-R-M) followed by two fully connected layer and then output through a softmax layer. We are not considering the softmax conversion as part of the neural network architecture. In the second, larger architecture, we increased the width of neural network now including 5 layers of the C-R-M sequence and two fully connected layers.

Because it was easier and quicker to train the shallow CNN, we used this for prototyping and hyperparameter tuning. On this architecture, we used different number of filters used in each convolution layer, different window size and strides and also tried different activations. Effect of activations was more apparent in the run time, this could be because of the implementations of the activation functions in TensorFlow. We have tried and tested relu, elu, sigmoid and tanh activation functions. Using sigmoid and tanh gave the worst results in terms of cost and accuracy while relu, elu activation units were much faster and more accurate for the same number of epochs. In all the tests, we used the Xavier initialization\cite{glorot2010understanding} method for all the weights. The convergence test followed the results form the original paper where elu was introduced by Clevert and others in \cite{clevert2015fast}.

With parameters obtained from these experiments, we trained the wider CNN. The reason behind doing so is quite obvious, the wider CNN takes a longer time to train and it is not feasible to search for hyperparameters in this setting. With the hope that the optimal hyperparameters of the shallow networks also worked for the deeper CNN because they follow the same architecture we begun our experiments. The experiments on the wider CNN were done in two folds.

Analogous to the number of hidden units  and layers in traditional densely connected neural networks, adding layers corresponded to finding hidden features of higher orders .Having more units in the same layer meant better feature extraction in the CNNs. Following this intuition, we have increased the number of filters used in each convolution layer. This increased the number of parameters and the training time. It eventually helped in getting better overall results which are explained in further sections

One interesting experiment we did for this setting was the use of dropout\cite{srivastava2014dropout} as a regularization method. We tried different usages of the additional dropout layer corresponding to its addition in various layers and combinations. This is again a search for an optimal hyperparameter (let's say keep probability). We again  tried this on the shallow CNN due to time constraints. The results though were counterintuitive. As we trained the model for higher training steps (epochs), we expected the model to overfit which was the case when we did not use dropout. Surprisingly, the results followed a similar pattern even with dropout but with a lower training and validation accuracy. There is some research\cite{gal2015bayesian,kendall2016modelling,krizhevsky2012imagenet} that uses dropout in CNNs and advices against using it.
\begin{figure}[h!]
\includegraphics[width = 0.9 \linewidth]{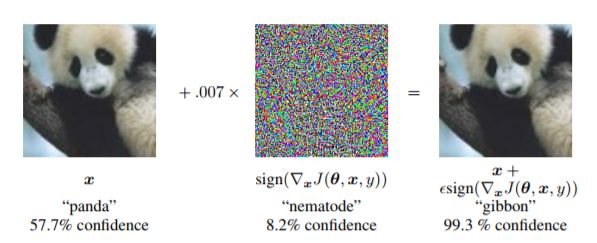}
\caption{A demonstration of fast adversarial example generation applied to GoogLeNet (Szegedy
et al., 2014a)\cite{szegedy} on ImageNet. By adding an imperceptibly small vector whose elements are equal to
the sign of the elements of the gradient of the cost function with respect to the input, we can change GoogLeNet’s classification of the image. Image:\cite{goodfellow2014explaining}}
\label{fig:from Goodfellow et al}
\end{figure}
\subsection{Virtual Adversarial Training}
In the Deep Learning book by Goodfellow et al,\cite{Goodfellow-et-al-2016}, several other regularization methods like image augmentation, noise addition, early stopping and virtual adversarial training\cite{Miyato,goodfellow2014explaining,szegedy} were mentioned. For our problem, because we converted the audio files into images and thereby converting a speech recognition problem into a viable image recognition one just to use CNN, we could use most of the usual regularization methods in CNN domain. Among these, we decided to try Adversarial Training as a regularization method for our problem. The idea behind this was that the dataset originally contained audio samples of 31 different classes but within each class the audio samples seem to follow similar distributions (this is more natural too, think of how peoples voice differ \cite{muda2010voice}, this could be learned as one hidden feature in the image domain).

 By adding small noise to the original distribution, we could potentially generate images that are "visually similar" but are actually different and train the model on a mixture of original and noisy images with same labels so that the model generalizes to these small disturbances. Although we successfully generated visually similar images, we do not actually know the effect of this transformation in the audio domain.

Following the ideas in \cite{Miyato,goodfellow2014explaining,szegedy}, we augmented the dataset with interesting changes. First we used OpenCV library in Python to convert the images (obtained from converting audio files) into Numpy arrays and then normalized the pixel intensities (divided by 255). We then added a perturbed version (Numpy array) of the signum output of the matrix created in the previous step to itself. We also created a new matrix whose entries $\hat{X}_{i,j,k,l} = X_{i,j,k,l} + 0.001*std(X)$ and these two generated noisy inputs were concatenated with the training data drawn from the original matrix along with corresponding labels. When trained with the same model on this with same hyperparameters, the model has shown faster convergence (in terms of epochs) as can be seen figure 9. This could be because of generalization (regularization) or because the model has seen thrice as much data.
\begin{figure}[h!]
\includegraphics[width = 0.9\linewidth]{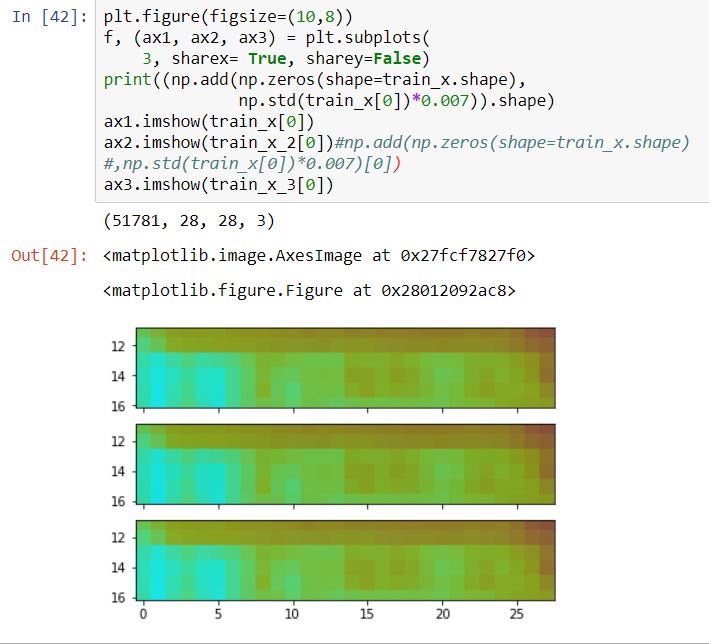}
\caption{First image is the original example, second is the sign perturbed image and the last one is the standard deviation perturbed image}
\label{fig: Adversarial Examples vs the original example}
\end{figure}
So we sampled equal amount of training data from this augmented dataset and carried out results. We achieved slightly better accuracies on both training and validation parts. This is significant because using dropout gave around 53\% accuracy while the vanilla CNN and the CNN with adversarial training gave more than 88\% accuracy. This shows that Virtual Adversarial Training (VAT) is a better regularization method, at least for this setting.

\section{Experiment Set Up}

\subsection{Dataset}
The Speech Commands Dataset \cite{speechcommands} which was used for this project is freely available. The data consists of 65,000 labeled audio clips, sampled from thousands of different speakers from all over the world, and each recording is of different quality. Each training example is a one-second-long audio file where a voice speaks one English word. While there are a total of 30 different possible words that can appear in a training example, our model will identify ten particular words, corresponding to the most basic voice commands:  \textit{YES}, \textit{NO}, \textit{UP}, \textit{DOWN}, \textit{LEFT}, \textit{RIGHT}, \textit{ON}, \textit{OFF}, \textit{STOP}, and \textit{GO}. All other words will receive a label of \textit{UNKNOWN}. Additionally, the dataset contains audio examples containing only noise, which our model must identify as \textit{SILENCE}. 

80 percent of the data was used for training, and 20 percent for validation. The raw audio data has a sample rate of \SI{16}{kHz} and is one second long, resulting in 16000-item vectors. 
Sections of the project which use spectrograms use images resampled to a size of 28x28 grayscale pixels, while amplitude-vs.-time plots were 100x100 pixels. 

\subsection{Framework}
The package TensorFlow was used to build our models,
chosen over competing libraries and in preference to higher-level approaches such as Keras because of the increased flexibility afforded by a lower-level framework, combined with the simple fact of our greater familiarity with TensorFlow.

\section{Results and Discussion}
\subsection{CNNs and Virtual Adversarial Training}

\begin{figure}[h!]
\includegraphics[width = 0.9\linewidth]{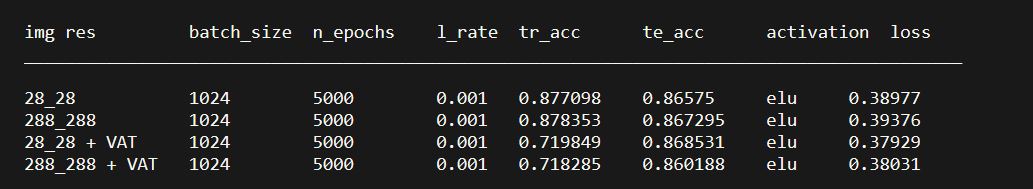}
\caption{Some results using less filters on wide CNN }
\end{figure}
In figure 8 we show the results on two sets of datasets 28x28 corresponds to the pixel size 28x28 and 288x288 pixel size. It did not help to use a higher resolution image. In this table, the results correspond to a run where the model was allowed to stop training when it hits a cost threshold shown as "thres" in the image. The "VAT" indicates that the model was trained with adversarial inputs. It is evident that during the VAT training, convergence occurred much sooner than the corresponding vanilla (column n\_epochs with (e) indicates exit/ending epoch). We also show some results obtained on the deeper CNN but with less number of filters.

\begin{figure}
\includegraphics[width = 0.9 \linewidth]{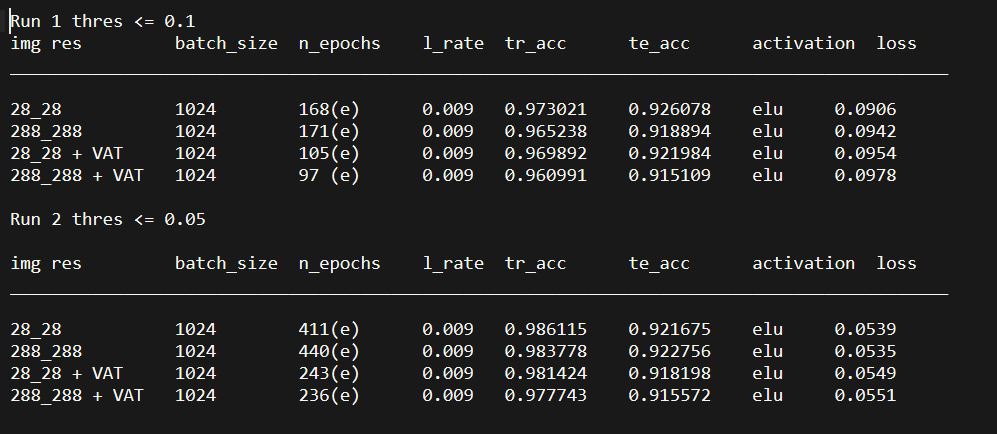}
\caption{Adversarial Training Results and Comparison with Vanilla CNN}
\end{figure}

\begin{figure}
\includegraphics[width = 0.9\linewidth]{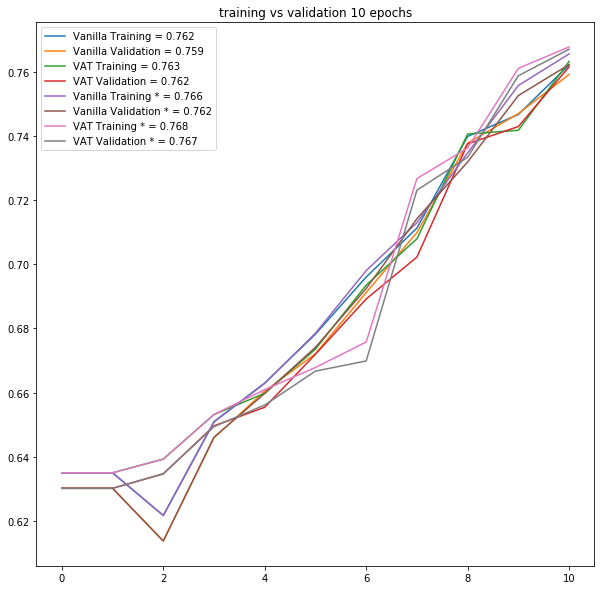}
\caption{\label{fig:10epa}
	increase in training and validation accuracy 
	over the first ten epochs for the low-latency convolution
    and VAT models: a zoomed-in version of \autoref{fig:500epa}}
\end{figure}

\begin{figure}[h!]
\includegraphics[width = 0.9\linewidth]{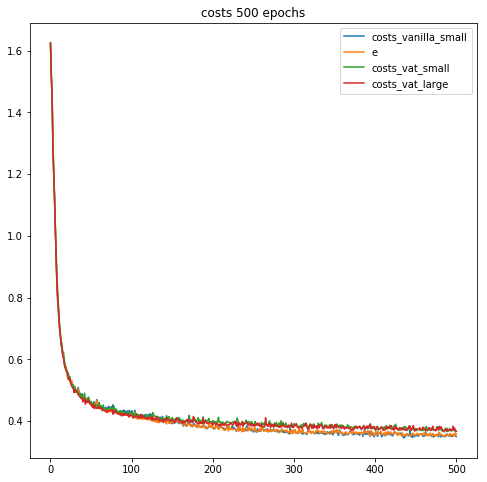}
\caption{\label{fig:500epc}
	decrease of costs over 500 epochs for the low-latency
	convolution and VAT models: a zoomed-out version of
    \autoref{fig:10epc}}
\end{figure}
In figures 10 and 11 we show the results in terms of training / validation accuracies vs epochs of the models that we trained and tested. The second image (figure 11) shows the behavior of the models in the first 10 epochs. It can be seen that the accuracies are much closer and are scaling at a similar rate when we use adversarial Training. 

\begin{figure}[h!]
\includegraphics[width = 0.9\linewidth]{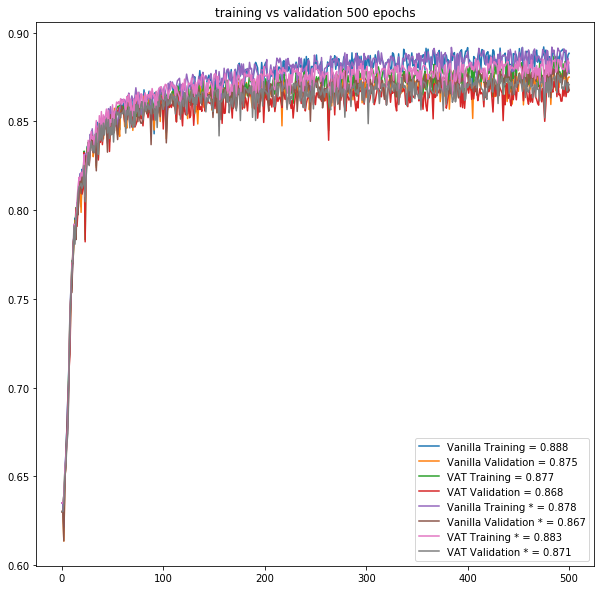}
\caption{\label{fig:500epa}
	increase of training and validation accuracy over 500 
    epochs for the low-latency convolution and VAT models:
    a zoomed-out version of \autoref{fig:10epa}}
\end{figure}
In figure 12 we can see the behavior of loss function with respect to time for all the models. Given the model capacity and training data, the effect of regularization induced by VAT is not entirely clear from the cost vs epochs graph. Although, it can be seen that the cost function with larger VAT model has converged quickly after approximately 120-150 epochs whereas the other models appear to have converged a few tens of epochs later.

\begin{figure}[h!]
\includegraphics[width = 0.9\linewidth]{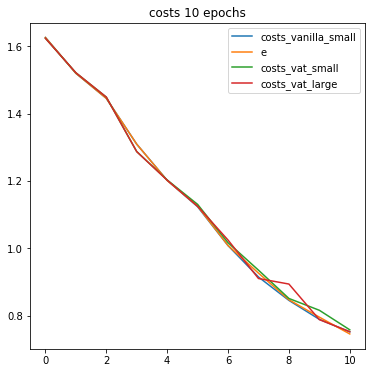}
\caption{\label{fig:10epc}
	reduction of cost over the first ten epochs 
	for the low-latency convolution and VAT models:
    a zoomed-in version of \autoref{fig:500epc}}
\end{figure}
The behavior of cost function in the first 10 epochs is seen image 13.
\subsection{Tensorflow MNIST and Basic CNN}

We also tried the CNN architecture described in the MNIST Tutorial by Tensorflow\cite{mnistexpert} on two different conversions of our training data to image format. When trained on images of the amplitude-vs.-time graph of the audio file, the network attained a validation accuracy of 50.3\%, compared to 52.5\% on log spectrogram images, suggesting that our intuition was correct regarding the most informative input formats.

\subsection{Low-latency Convolutional Model}
Beginning with the basic low-latency convolutional model described above, we tried varying a number of different hyperparameters and comparing results.
\begin{figure}[h!]
	\centering
    \frame{\includegraphics[width=0.7\linewidth]{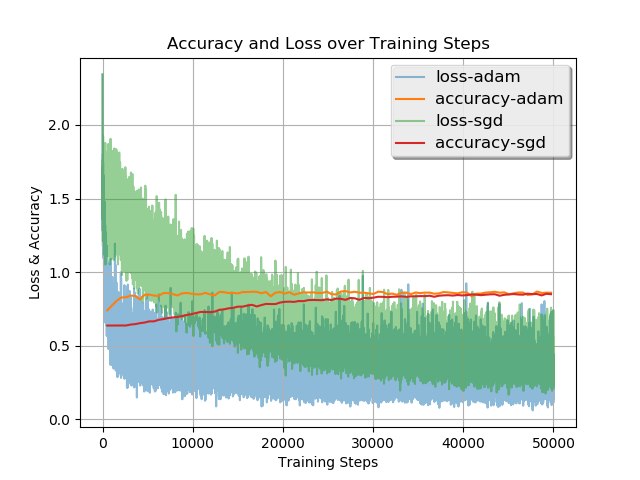}}
    \caption{\label{fig:optimizer_results}
    the evolution of training cross-entropy loss (blue and green)
    and validation accuracy (red and orange) 
    compared between Adam and SGD optimization;
    Adam converges faster than SGD but reaches the same results}
\end{figure}

\paragraph{Optimization Algorithm}
Because our optimization problem is nonconvex, choices such as the initialization and the precise method of optimization used begin to matter. With this in mind, we compared two different approaches to gradient descent: the adaptive ADAM optimization algorithm vs.\  a typical stochastic gradient descent.
\autoref{fig:optimizer_results} compares the evolution of both loss and accuracy over time between Adam and SGD. By both metrics, Adam converges noticeably faster without substantially changing the final point of convergence. 

\begin{figure}[h!]
	\centering
    \frame{\includegraphics[width=0.7\linewidth]{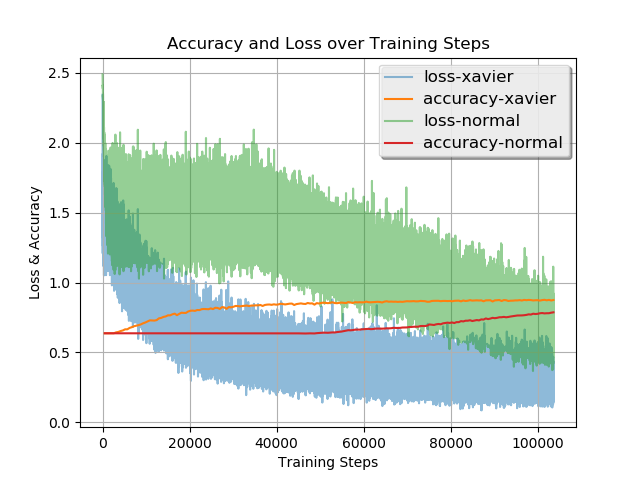}}
    \caption{\label{fig:initialization_results}
    the evolution of training cross-entropy loss (blue and green)
    and validation accuracy (red and orange) compared between 
    Xavier and truncated normal initialization; Xavier converges
    much faster and may attain better results}
\end{figure}

\paragraph{Initialization}
The nonconvexity of our problem also led us to consider alternate initialization methods. 
We compared the Xavier random initialization method (detailed elsewhere, but in brief, values are initialized uniformly within a small interval) to the \textit{de facto} standard truncated normally-distributed initialization. 
\autoref{fig:initialization_results} compares the evolution of both loss and accuracy over time between models differing only in the choice between Xavier and truncated normal initialization.
In this case even more so than in the case of the choice of optimization algorithm, there is a marked difference in the rate of convergence, but here there may also be a difference in the final result.
It is not clear whether the network with truncated normal initialization has fully converged by the end of the training, while the network with Xavier initialization has spent the majority of the same amount of time hovering around the same loss and accuracy, i.e. already converged. 
Furthermore, by the end of this trial, the accuracy and loss of the network with truncated normal initialization are markedly worse than those of the network with Xavier initialization.

\paragraph{Spectrogram Parameters}
Because we converted the problem of audio recognition to the image domain using spectrograms, the input stage of the pipeline also provided a number of different parameters to tune. 
A spectrogram is generated by taking subsequences of a certain fixed length from the input sequence, multiplying them by a windowing function\footnote{a regularization method without which spurious high-frequency components will invariably be created in the output, obscuring more important parts of the signal}, taking the discrete Fourier transform of each windowed subsequence, and pasting them together in the form of an image. 
The way those subsequences are chosen does make a difference to the final results. 
We tried varying three different parameters of the spectrogram: the number of buckets used for frequency counting, the size of the window, and the spacing between different windows.
The plots for all of these parameters specify the number of 

\begin{figure}[h!]
	\centering
    \frame{\includegraphics[width=0.7\linewidth]{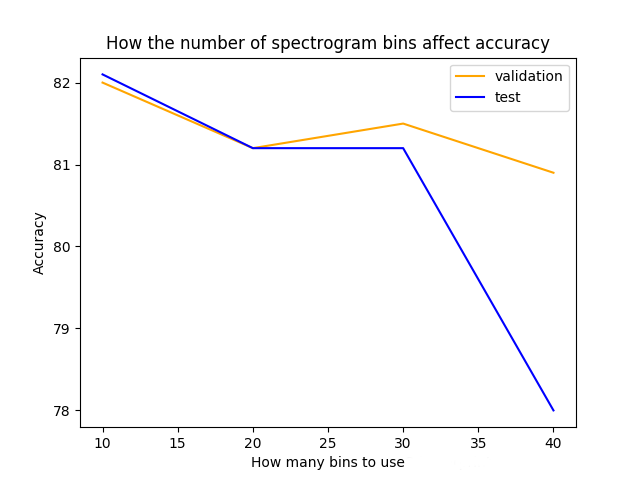}}
    \caption{\label{fig:dct_coeff_vs_accuracy}
    effect of the number of frequency-counting buckets on the
   	accuracy of the low-latency convolution model.
    The model did not benefit from the increase in available
    data caused by increasing the number of buckets.}
\end{figure}
One might expect that increasing the number of buckets in the frequency counting (that is, the number of sections into which the frequency axis was divided, or the pixel height of the spectrogram) would help the network discriminate better due to the increase in the amount of data contained in each training example, but in fact the effect was the opposite: across the whole range of values we tried, from 10 to 40 frequency buckets, increasing the number of buckets only worsened the performance of the network. 
This can be seen in \autoref{fig:dct_coeff_vs_accuracy}.

The stride between windows was found to have little effect within a certain range. When the stride was too low or too high, however, results began to worsen. 
This is shown in \autoref{fig:window_stride_vs_accuracy}.
These results are to be expected, because excessively low stride means that the windows overlap significantly, creating a large amount of redundant data which the network must sort through.
Likewise, excessively high stride naturally results in missing out on important parts of the data when certain components happen to fall in the gaps between the windows.  
This could cause certain features to not show up in the spectrogram at all, in which case it is of course impossible for the network to learn them.
\begin{figure}[h!]
	\centering
    \frame{\includegraphics[width=0.7\linewidth]{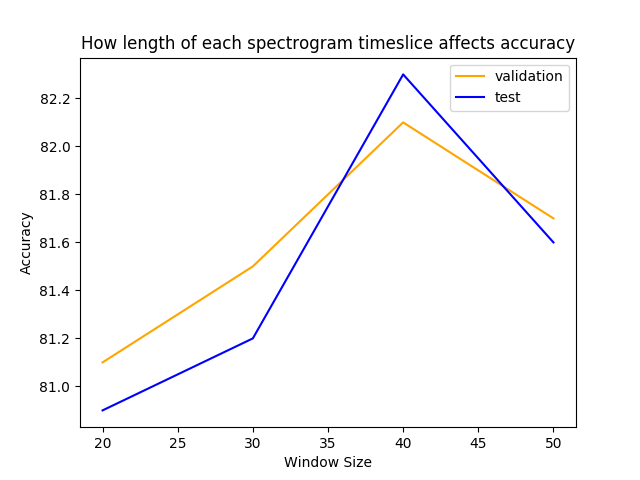}}
    \caption{\label{fig:window_size_vs_accuracy}
    effect of the spectrogram window size on the accuracy of 
    the low-latency convolution model.
    There is a local optimum, as there was for stride in
    \autoref{fig:window_stride_vs_accuracy}.} 
\end{figure}
Like the stride length, the size of the convolutional window had an optimum value about which the final accuracy fell off noticeably.
This behavior is shown in \autoref{fig:window_size_vs_accuracy}.
One would expect that the only effect of changing the window size would be that the meaning of the stride value changes: since stride remains fixed, a large window will overlap substantially with adjacent windows, while a small window will leave gaps. 
So the entire observed variation in final accuracy with window size can likely be attributed to the same reasons as  the variation in accuracy with stride.
\begin{figure}[h!]
	\centering
    \frame{\includegraphics[width=0.7\linewidth]{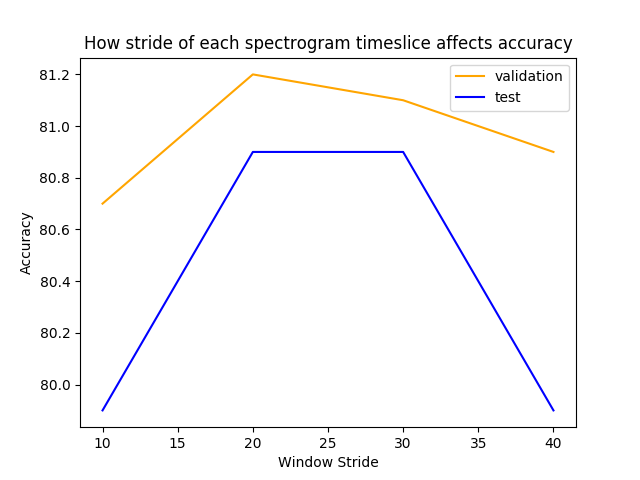}}
    \caption{\label{fig:window_stride_vs_accuracy}
    effect of the spectrogram window stride on the accuracy
    of the low-latency convolution model.
    For low values of stride, there is too much redundancy,
    while larger values result in lost information.}
\end{figure}

\paragraph{Noise}

In addition to varying hyperparameters, we also tried adding noise to the audio signal prior to computing the spectrogram. 
This is crucial for real-world applications because the desired target words will never be spoken completely in the absence of competing stimuli. There will always be background noise from many sources and in many forms, such as the pure white noise generated by the thermal voltage in the analog microphone amplifier, the unordered crashes of nearby remodeling, or the highly-ordered distraction of an irrelevant conversation. 
\begin{figure}[h!]
	\centering
    \frame{\includegraphics[width=0.7\linewidth]{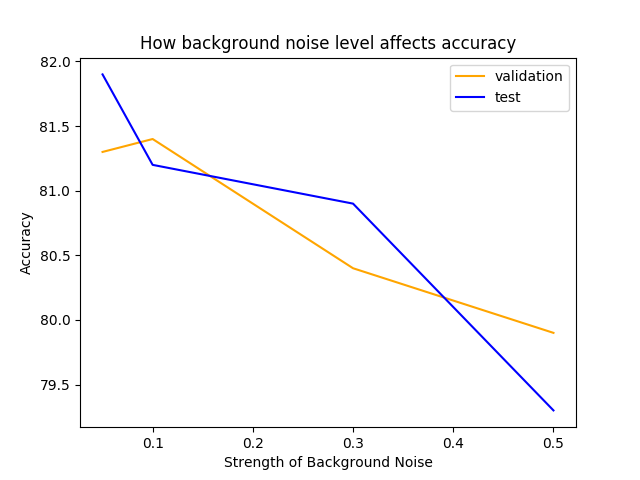}}
    \caption{\label{fig:noise_vs_accuracy}
    effect of added background noise on the final accuracy of
    the low-latency convolution model.
    The horizontal axis is signal-noise ratio in linear units.}
\end{figure}
Our sample dataset included samples of the audio interference from various sources of noise, such a dishwasher, which could be mixed in at random to simulate background noise.
We tried this at various different noise-to-signal ratios over a range from 0 to 0.5 (corresponding to signal-noise-ratios of infinity up to \SI{6}{dB}), with results as shown in \autoref{fig:noise_vs_accuracy}. 
Of course, increased background noise did degrade the performance of the network, but the effect was comparatively slight: on the order of a few percentage points accuracy. 
This suggests that this approach is fairly robust to noise; even humans' accuracy at comparable tasks is noticeably degraded by low signal-to-noise ratios.

\section*{Conclusion}
In this project we tackled the speech recognition problem by applying different CNN models on image data formed using log spectrograms of the audio clips. We also successfully implemented a regularization method "Virtual Adversarial Training" that achieved a maximum of 92\% validation accuracy on 20\% random sample of the input data.

The significant work done in this project was the demonstration of how to convert a problem in audio recognition into the better-studied domain of image classification, where the powerful techniques of convolutional neural networks are fully developed. 
We also saw, particularly in the case of the low-latency convolution model, how crucial good hyperparameter tuning is to the accuracy of the model. A great number of hyperparameters must be tuned, including the many choices that go into network architecture, and any of the hyperparameters, poorly chosen, can make or break the overall performance of the model.
Another contribution was the use of adversarial training to provide a regularization effect in audio recognition; this technique improved results relative even to well-established techniques such as dropout, and therefore has promising applications in the future.

\break

\bibliographystyle{ACM-Reference-Format}
\bibliography{mybib} 

\end{document}